\documentclass[letterpaper]{article} % DO NOT CHANGE THIS
\usepackage{aaai2026}  % DO NOT CHANGE THIS
\usepackage{times}  % DO NOT CHANGE THIS
\usepackage{helvet}  % DO NOT CHANGE THIS
\usepackage{courier}  % DO NOT CHANGE THIS
\usepackage[hyphens]{url}  % DO NOT CHANGE THIS
\usepackage{graphicx} % DO NOT CHANGE THIS
\usepackage{natbib}  % DO NOT CHANGE THIS AND DO NOT ADD ANY OPTIONS TO IT
\usepackage{caption} % DO NOT CHANGE THIS AND DO NOT ADD ANY OPTIONS TO IT
\usepackage{algorithm}
\usepackage{algorithmic}

\usepackage{newfloat}
\usepackage{listings}
\DeclareCaptionStyle{ruled}{labelfont=normalfont,labelsep=colon,strut=off} % DO NOT CHANGE THIS
\floatstyle{ruled}
\newfloat{listing}{tb}{lst}{}
\floatname{listing}{Listing}
\usepackage{multirow}
\usepackage{bbding}
\usepackage{booktabs}
\usepackage{amsmath}
\usepackage{amssymb}

\title{SAM-DAQ: Segment Anything Model with Depth-guided Adaptive Queries for RGB-D Video Salient Object Detection}
\author{
    Jia Lin\textsuperscript{\rm 1},
    Xiaofei Zhou\textsuperscript{\rm 1}\equalCorresponding,
    Jiyuan Liu\textsuperscript{\rm 1},
    Runmin Cong\textsuperscript{\rm 2},
    Guodao Zhang\textsuperscript{\rm 1},
    Zhi Liu\textsuperscript{\rm 3}\equalCorresponding,
    Jiyong Zhang\textsuperscript{\rm 1}
}
\affiliations{
    \textsuperscript{\rm 1}Hangzhou Dianzi University, Hangzhou, China\\
    \textsuperscript{\rm 2}Shandong University, Jinan, China\\
    \textsuperscript{\rm 3}Shanghai University, Shanghai, China\\

    lin\_j@hdu.edu.cn, zxforchid@outlook.com, hankliu@hdu.edu.cn, rmcong@sdu.edu.cn,\\
    guodaozhang@hdu.edu.cn, liuzhisjtu@163.com, jzhang@hdu.edu.cn
}

\usepackage{bibentry}
\begin{document}

\maketitle

\begin{abstract}
% SAM
Recently segment anything model (SAM) has attracted widespread concerns, and it is often treated as a vision foundation model for universal segmentation. Some researchers have attempted to directly apply the foundation model to the RGB-D video salient object detection (RGB-D VSOD) task, which often encounters three challenges, including the dependence on manual prompts, the high memory consumption of sequential adapters, and the computational burden of memory attention.
% 我们的方法
To address the limitations, we propose a novel method, namely \textbf{S}egment \textbf{A}nything \textbf{M}odel with \textbf{D}epth-guided \textbf{A}daptive \textbf{Q}ueries (SAM-DAQ), which adapts SAM2 to pop-out salient objects from videos by seamlessly integrating depth and temporal cues within a unified framework. 
% PAMIE
Firstly, we deploy a parallel adapter-based multi-modal image encoder (PAMIE), which incorporates several depth-guided parallel adapters (DPAs) in a skip-connection way. Remarkably, we fine-tune the frozen SAM encoder under prompt-free conditions, where the DPA utilizes depth cues to facilitate the fusion of multi-modal features.
% QTM
Secondly, we deploy a query-driven temporal memory (QTM) module, which unifies the memory bank and prompt embeddings into a learnable pipeline. Concretely, by leveraging both frame-level queries and video-level queries simultaneously, the QTM module selectively extracts temporal consistency features, iteratively updates the temporal representations of the queries.
% 实验
Extensive experiments are conducted on three RGB-D VSOD datasets, and the results show that the proposed SAM-DAQ consistently outperforms state-of-the-art methods in terms of all evaluation metrics.
\end{abstract}

\begin{links}
    \link{Code}{https://github.com/LinJ0866/SAM-DAQ}
    % \link{Datasets}{https://aaai.org/example/datasets}
    % \link{Extended version}{https://aaai.org/example/extended-version}
\end{links}

\section{Introduction}

\begin{figure*}[h]
  \centering
  \includegraphics[width=\linewidth]{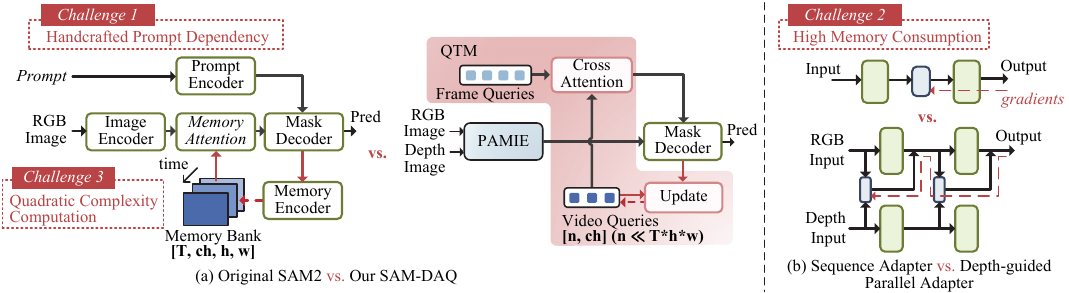}
  \caption{High-level illustration of our SAM-DAQ.}
  \label{fig:teaser}
\end{figure*}

% 第一段，介绍SOD任务，其不足引出下一段多模态
% sod~\cite{ke2023segment, hui2024endow, xiong2024sam2, li2025kansam} 
Salient object detection (SOD)~\cite{borji2019salient} aims to identify and highlight the most attractive objects in visible images, which has been widely applied to many related areas, such as object tracking~\cite{zhang2020non, zhou2021saliency}, robotics~\cite{bao2025ifenet, huang2025multi} and image retrieval~\cite{fan2015improving}.
% 单模态任务在困难场景中的表现有限
%Despite promising advances, 
However, the conventional SOD methods relying solely on RGB data often present unsatisfactory performance when dealing with challenging scenarios, such as cluttered background, occlusions, and low light conditions. %, resulting in notable performance degradation.

% 第二段，多模态提供了很好的解决方案
To overcome the aforementioned limitations, some researchers have increasingly explored the introduction of multiple modalities.
% depth 
RGB-D SOD methods~\cite{qu2017rgbd, han2017cnns, cong2023point, bao2024quality} leverage depth information to provide robust spatial structure, effectively mitigating background distractions.
% temporal
Meanwhile, video SOD (VSOD) methods~\cite{wang2017video, li2018flow, liu2023part, zhou2023sti} incorporate temporal cues into the SOD task, where they major in digging an effective characterization of motion cues. 
Embarking on this, researchers attempt to introduce both depth and temporal information to the SOD task, namely RGB-D VSOD.

% 第四段，介绍SAM，及其问题：1）需要提供prompt；2）sequence adapter带来的显存消耗大；3）memory attention计算量大
% sam
Recently, the segment anything model (SAM)~\cite{kirillov2023segment}, which is treated as a vision foundation model for universal segmentation, has attracted more and more concerns.
% SAM2
Embarking on SAM, SAM2~\cite{ravi2024sam} further extends its capability to video segmentation by incorporating a memory mechanism, which captures inter-frame dependencies via a memory bank.
% adapt SAM的挑战
However, directly applying SAM2 to the RGB-D VSOD task introduces several challenges.
% 1）SAM需要手工输入prompt
Firstly, to guide the object segmentation task, SAM2 requires manual prompts (e.g., points, boxes and masks), but the RGB-D VSOD task cannot provide such information during the inference stage (Fig.~\ref{fig:teaser}(a)).
% 2）adapter，然而，由于反向传播需要保存梯度，显存消耗大
Secondly, sequential adapters~\cite{houlsby2019parameter} used for prompt-free adaptation cause high GPU memory consumption during training, because the backward gradients must traverse the entire encoder~\cite{diao2024unipt} (Fig.~\ref{fig:teaser} (b)).
% 3）memory attention计算量大
Finally, the memory attention mechanism in SAM2 imposes a high computational burden due to the extensive correlation computations between the current-frame features and the large memory bank.

% 第五段：针对上述问题，提出我们
To overcome the aforementioned limitations, we propose a Segment Anything Model with Depth-guided Adaptive Queries (SAM-DAQ), which adapts SAM2 for the RGB-D VSOD by seamlessly integrating depth and temporal cues.
% PAMIE。重点在skip-connected结构与depth指导RGB
Specifically, we propose a parallel adapter-based multi-modal image encoder (PAMIE) that incorporates a series of depth-guided parallel adapters (DPAs).
This design enables efficient fine-tuning of the frozen vision foundation model-based encoder under prompt-free conditions, while leveraging depth information as guidance to facilitate the fusion of RGB and depth features.
% QTM
In addition, we develop a query-driven temporal memory (QTM) module, which replaces the memory bank and prompt embedding with learnable queries (\emph{i.e.}, object queries~\cite{carion2020end}), to capture temporal dependencies among different frames. Particularly, by leveraging both frame-level queries and video-level queries, the QTM module can selectively highlight visually attractive regions by incorporating the temporal consistency and iteratively update the temporal characterization of the queries. In this way, we can obtain effective learnable embeddings for the mask decoder.
% contribution
In summary, our contributions can be summarized as follows:

\begin{itemize}
    \item We propose a Segment Anything Model with Depth-guided Adaptive Queries (SAM-DAQ) to conduct RGB-D VSOD, which effectively adapts the vision foundation model by deploying a parallel adapter-based multi-modal image encoder (PAMIE) and a query-driven temporal memory (QTM) module.
    \item We propose a PAMIE to enable prompt-free fine-tuning with minimal memory consumption and facilitate effective RGB-D fusion by using depth-guided parallel adapters (DPAs).
    \item We propose a QTM module to unify the memory bank and prompt embeddings into a learnable pipeline by using frame-level queries and video-level queries, where the former extracts visually attractive information from each frame and the latter captures temporal dependencies among different frames via cross-attention.
   \item Extensive experiments on three RGB-D VSOD datasets firmly demonstrate the superiority of the proposed SAM-DAQ over the state-of-the-art models.
\end{itemize}

\section{Related Work}

% 参考Salient Object Detection in RGB-D Videos，分为early, late and middle fusion三个阶段
\subsection{RGB-D Salient Object Detection}
Depth information provides spatial cues resistant to contextual distractions and is widely as regarded a critical complement for RGB information. Existing RGB-D SOD methods typically follow three fusion paradigms~\cite{mou2024salient}, namely early fusion, middle fusion and late fusion.
% early fusion，将输入按通道拼接
The early fusion strategy~\cite{qu2017rgbd} treats the depth map as an additional channel to the RGB image, and directly concatenates the two modal images to form a 4D input.
% late fusion，两个模态单独送入网络推理，融合输出的结果（再差）
The late fusion strategy~\cite{han2017cnns} processes the two modalities independently, which generates the final mask by fusing their coarse predictions. However, the interaction between the modalities occurs at the output only, which limits the fusion of the two modal features. 
% middle fusion
The middle fusion strategy~\cite{cong2023point, bao2024quality} is widely employed to capture and exploit multi-modal correlations. 
% CNN->Transformer
For instance, Cong \emph{et al.}~\cite{cong2023point} combine the multi-modal CNN features and then utilize the fused features to refine the transformer decoder. 
However, prior studies have revealed inherent limitations in the depth modality, including a lack of informative content~\cite{hao2024primkd} and unstable quality~\cite{bao2024quality}.

% 分为传统方法（时间窗、光流法）, memory bank和query-based（使用learnable queries作为存储结构的） 
\subsection{Video Salient Object Detection}
Video SOD (VSOD) extends SOD by leveraging temporal information across frames to ensure spatial-temporal consistency.
% traditional
Early works~\cite{wang2017video, li2018flow, ji2020casnet} utilize optical flow and attention mechanisms to capture motion cues, but their accuracy heavily depends on flow quality and motion scale~\cite{singh2024dsfnet}.
% memory
To better model long-term dependencies, memory-based frameworks~\cite{oh2019video, cheng2022xmem} encode the previous frame's features and their predictions, and store the generated representations in a memory bank.
However, interacting with a large memory bank can introduce substantial computational consumption.
% query-based
To tackle the issue, query-based methods~\cite{fang2024learning, wang2023look} utilize the learnable query to focus on relevant features selectively, achieving both high accuracy and efficiency.

Recent RGB-D VSOD methods aim to integrate depth and temporal cues simultaneously.
For instance, Li~\emph{et al.}~\cite{li2023dvsod} store previously fused RGB-D features into memory, extending memory networks to RGB-D settings.
Mou~\emph{et al.}~\cite{mou2024salient} fuse flow and multi-modal features via holistic multi-modal attentive paths (HMAPs).
Lin \emph{et al.}~\cite{lin2024vidsod} treat RGB, depth, and optical flow equally, and deploy intermediate supervision on their respective encoders. % to promote the detection performance.
In addition, Suolang \emph{et al.}~\cite{suolang2025lightweight} propose a lightweight cross-shift module that efficiently fuses auxiliary depth and temporal cues.

\subsection{Segment Anything Model}

The Segment Anything Model (SAM)~\cite{kirillov2023segment, ravi2024sam} is a vision foundation model trained on large-scale data for universal image segmentation. While it generalizes well, its reliance on manual prompts (e.g., points, boxes), making it impractical for the video salient object detection task.

% 利用传统网络的输出结果作为coarse mask，生成prompt送入SAM 和 只取encoder
To eliminate the need for manual prompts, several works~\cite{zhang2024uv, ayzenberg2024protosam, xie2025rfmedsam} attempt to generate pseudo-prompts from coarse masks, while others~\cite{yang2024sam, xiong2024sam2, xu2025dgsunet} only use SAM's encoder for feature extraction.
% 上述方法并未改进SAM本身范式
Generally, the above methods tailor SAM for specific tasks without modifying its intrinsic architecture.
Recent methods have explored parameter-efficient fine-tuning (PEFT) strategies.
% Wang adapter
%For instance,
Wang \emph{et al.}~\cite{wang2024adapting} insert adapter~\cite{houlsby2019parameter} between encoder blocks, % to fine-tune the SAM encoder.
% Zhong，LoRA（插入至transformer模块中）
while Zhong \emph{et al.}~\cite{zhong2024convolution} integrate LoRA~\cite{hu2022lora} into transformer layers.
% 梯度传到问题
However, these sequential adapter structures tend to incur high memory consumption, as backward gradients must propagate through the entire encoder~\cite{diao2024unipt}.
% %parallel
% Different from the aforementioned efforts, we propose a parallel adapter-based multi-modal encoder (PAMIE), which employs skip-connected depth-guided adapters to reduce training memory and improve RGB-D fusion. 

To extend SAM to the video segmentation task, Yue \emph{et al.}~\cite{yue2024sam} propose a flow reconstruction technique to guide SAM in object discovery. Deng \emph{et al.}~\cite{deng2024memsam} present three distinct types of memory banks to mitigate the adverse effects of speckle noise and motion artifacts during memory prompting. To the best of our knowledge, no previous work has attempted to encode memory queries within the video SAM-based framework.
% % Ours
% Different from the aforementioned efforts, we introduce query-based temporal memory (QTM) module that unifies prompt representation and temporal modeling.
% Specifically, to fully leverage the power of learnable queries, we integrate the memory bank and prompt embeddings into a unified learnable pipeline via a query-driven temporal memory (QTM) module. Specifically, we design frame-level queries to extract salient spatial features and video-level queries to capture inter-frame dependencies through iterative updates.

\section{Method}

\begin{figure*}[h]
  \centering
  \includegraphics[width=0.94\linewidth]{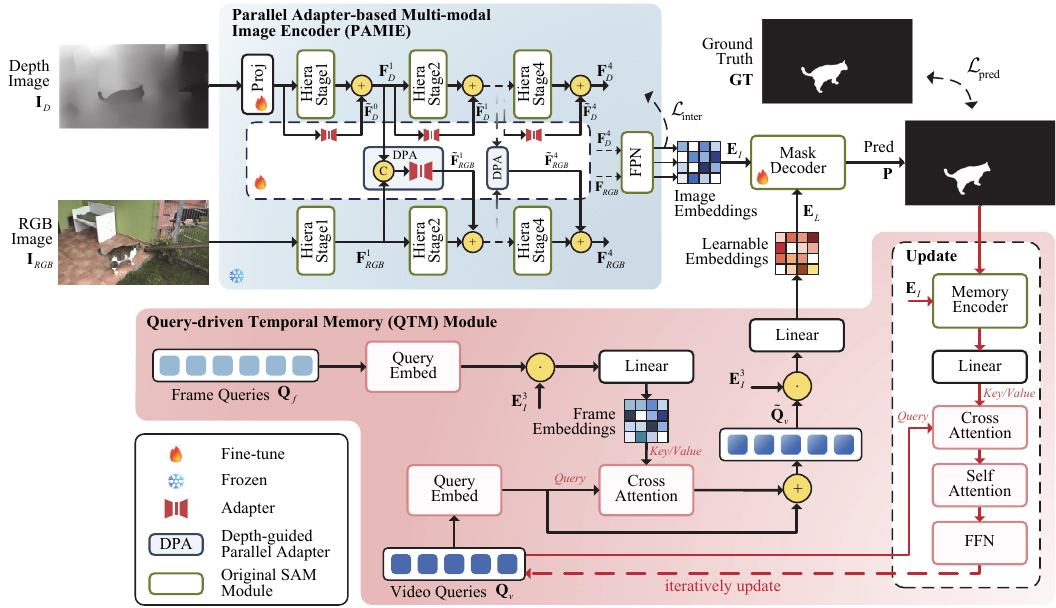}
  \caption{The overall architecture of the proposed Segment Anything Model with Depth-guided Adaptive Queries (SAM-DAQ) of a single frame.}
  \label{fig:model_structure}
\end{figure*}

% 介绍整体流程
\subsection{Overview}
As shown in Fig.~\ref{fig:model_structure}, we introduce the vision foundation model and propose a novel Segment Anything Model with Depth-guided Adaptive Queries (SAM-DAQ), which organically integrates the depth and temporal cues. Our SAM-DAQ consists of three components, namely a parallel adapter-based multi-modal image encoder (PAMIE), a query-driven temporal memory (QTM) module, and the mask decoder retained from the original SAM2.
% 每次处理一帧，为了便于阐述，我们仅介绍其中一帧的处理方式
%The input of the proposed SAM-DAQ is an image pair, namely RGB image and depth map.
Here, for simplicity, we present our method by detailing the process of a single frame. %Considering each input video of length $T$ frames is processed one frame at a time by SAM-DAQ, for readability, we illustrate our method by describing the processing of a single frame.
% 1st, PAMIE
Firstly, RGB image and depth image constitute an image pair $\{ \mathbf{I}_{RGB}, \mathbf{I}_{D} \}$, which is fed into PAMIE. Specifically, at each level, the PAMIE employs a depth-guided parallel adapter (DPA) to efficiently fine-tune the encoder under prompt-free conditions and fuse RGB and depth features. The encoder produces three-level image embeddings, namely $\mathbf{E}_I=\{ \mathbf{E}_I^i \}_{i=2}^4$.
% 2nd, QTM
Then, in the query-driven temporal memory (QTM) module, the highest-level image embeddings $\mathbf{E}_I^4$ progressively interact with static frame-level queries and iteratively update video-level queries, generating learnable embeddings $\mathbf{E}_L$.
% 3rd, mask decoder
Next, by integrating both the image embeddings and the learnable embeddings, the mask decoder predicts a high-quality segmentation result $\mathbf{P}$ for the current frame.
% 4th, memory update
Finally, the update mechanism in the QTM module leverages the current image embeddings and the corresponding high-quality segmentation result to update the video-level queries, ensuring that temporal dependencies are effectively captured. % for future prediction.

% adapter部分
\subsection{Parallel Adapter-based Multi-modal Image Encoder}
To sufficiently leverage the powerful segmentation performance and generalization of the vision foundation model, we propose a parallel adapter-based multi-modal image encoder (PAMIE) that parallelly integrates the original image encoder with the lightweight adapters.

% 送入RGB、Depth pairs，输出一个三 层的融合双模态信息的特征
Given the multi-modal image pair $\{ \mathbf{I}_{RGB}, \mathbf{I}_{D} \}$, % namely the current video frame, 
the input is processed separately by the Hiera~\cite{ryali2023hiera}.
% 具体的，考虑模态差异，将depth图像送入encoder前先行送入Proj，将两个模态表达投射至相同空间（文献支持：A Multimodal Translation-Based Approach for Knowledge Graph Representation Learning）
To address the inherent differences between RGB and depth modalities, we first use a linear depth projector~\cite{mousselly2018multimodal} to transform the depth input embeddings into the same feature space as the RGB modality.
% 冻结
After that, considering the limited RGB-D VSOD data and large number of trainable parameters in the SAM2 encoder, we freeze all encoder weights and employ parameter-efficient fine-tuning (PEFT) strategies to fine-tune the model.
% 我们的adapter
Specifically, unlike prior efforts~\cite{gao2024multi, wang2024adapting, zhong2024convolution}, which insert adapters sequentially between transformer blocks or apply LoRA to the query and value projections within each attention layer, we parallelly integrate the adapters in a skip-connection way.
% 特点2：省显存
This design not only facilitates the exploration of multi-modal information but also allows gradients to bypass the heavy transformer computations, significantly reducing memory consumption during the training stage.

% 我们设计了两种adapter结构。
Here, we design two kinds of parallel adapters.
% Depth分支
For the depth modality, the adapter connects the input and output of each Hiera block, which can be formulated as follows, 
% eq:adapter_depth，US、DS代表上采样与下采样参考自《Multi-Scale and Detail-Enhanced Segment Anything Model for Salient Object Detection》
\begin{equation}\label{eq:adapter_depth}
\left \{\begin{array}{l}
    \tilde{\mathbf{F}}_{D}^{i-1}=Adapter(\mathbf{F}_{D}^{i-1}) \vspace{1ex} \\
    \mathbf{F}_{D}^{i} = Hiera^{i}(\mathbf{F}_{D}^{i-1})+DS(\tilde{\mathbf{F}}_{D}^{i-1})
\end{array},
\right.
\end{equation}
where $Adapter$ denotes the adapter consisting of a down-projection linear layer followed by an activation function and an up-projection linear layer.
$Hiera^{i}$ represents the $i$-th Hiera block ($i = 1,2,3,4$), and $DS$ represents the bilinear downsampling operation. 
Here, $\mathbf{F}_{D}^{i}$ and $\tilde{\mathbf{F}}_{D}^{i}$ are the features of the $i$-th Hiera block and the corresponding adapter, respectively. Note that, $\mathbf{F}_{D}^{0}$ is the output of the depth projector.

% RGB分支，DPA
For the RGB modality, we utilize the depth-guided parallel adapter (DPA) to fuse RGB features and depth features. The whole process can be written as follows, 
% eq:adapter_RGB
\begin{equation}\label{eq:adapter_RGB}
\left \{\begin{array}{l}
    \tilde{\mathbf{F}}_{RGB}^{i-1}=Adapter(Cat(\mathbf{F}_{RGB}^{i-1}, \mathbf{F}_{D}^{i-1})) \vspace{1ex} \\
    \mathbf{F}_{RGB}^{i} = Hiera^{i}(\mathbf{F}_{RGB}^{i-1})+DS(\tilde{\mathbf{F}}_{RGB}^{i-1}) \vspace{1ex} \\
    \mathbf{F}_{RGB}^{1} = Hiera^{1}(\mathbf{F}_{RGB}^{0})
\end{array},
\right.
\end{equation}
where $Cat$ denotes concatenate operation and $\mathbf{F}_{RGB}^{i}$ is the RGB feature from the $i$-th Hiera block ($i = 2,3,4$). $\mathbf{F}_{RGB}^{0}$ means the input RGB image.
% 最终输出
Finally, after applying a feature pyramid network (FPN), PAMIE generates three-level image embeddings, namely $\mathbf{E}_I=\{ \mathbf{E}_I^i \}_{i=2}^4$.

% 监督
In addition, to further promote the fusion of the depth features and RGB features, we introduce a self-reasoning scheme that applies a lightweight convolution followed by a sigmoid activation function to each level image embedding $\mathbf{E}_I^i$, generating an intermediate prediction result, namely $\tilde{\mathbf{P}} = \{ \tilde{\mathbf{P}}^2, \tilde{\mathbf{P}}^3, \tilde{\mathbf{P}}^4 \}$.
% 我们的方案，只监督最高层
Notably, we only deploy the supervision to the highest-level image embeddings. %, meaning that $\tilde{\mathbf{P}} = \tilde{\mathbf{P}}^3$.

% 再次集中阐述该模块的作用（Multi-Scale and Detail-Enhanced Segment Anything Model for Salient Object Detection，Adapting Segment Anything Model to Multi-modal Salient Object Detection with Semantic Feature Fusion Guidance）

%Overall, the proposed PAMIE adapts the vision foundation model to the RGB-D SOD task, which is with few additional parameters and significantly reduces training memory requirements ().

% embedding生成部分，顺便简要介绍memory更新
\subsection{Query-Driven Temporal Memory Module}
% 目的：1）prompt需手工提供；2）memory bank运算复杂度
The manual prompts and the large memory bank restrict the application of the vision foundation model, especially in the VSOD task, where the former is difficult to acquire and the latter presents a high computational cost.
% Ours
To address the challenges, we propose a query-driven temporal memory (QTM) module, which unifies the prompt generation and the temporal modeling via a learnable query method.

% learnable queries，引出frame and video queries
%Inspired by~\cite{fang2024learning},
We introduce two sets of learnable queries, namely frame-level queries $\mathbf{Q}_f \in \mathbb{R}^{N_f \times c}$ and video-level queries $\mathbf{Q}_v \in \mathbb{R}^{N_v \times c}$, where $c$ is the hidden dimension, and $N_f$ and $N_v$ mean the number of frame-level queries and video-level queries, respectively.
% frame queries与image embeedings交互，生成frame embeddings（Adapting Segment Anything Model to Multi-modal Salient Object Detection with Semantic Feature Fusion Guidance）
As shown in Fig.~\ref{fig:model_structure}, the frame-level queries interact with the highest-level image embeddings $\mathbf{E}_I^4$, and in this way, we can acquire the saliency-related frame embeddings $\mathbf{E}_f$.
% video queries与frame embeddings
Besides, to incorporate temporal context, we perform cross-attention between video-level queries and frame embeddings. Here, both $\mathbf{Q}_{f}$ and $\mathbf{Q}_{v}$ are linearly projected via a query embedding layer, resulting in $\mathbf{Q}^{\prime}_{f}$ and $\mathbf{Q}^{\prime}_{v}$. The following operations can be defined as follows,
% eq:enhanced_video_queries
\begin{equation}\label{eq:enhanced_video_queries}
\left \{\begin{array}{l}
    \mathbf{E}_{f} = Linear(\mathbf{Q}^{\prime}_{f} \cdot \mathbf{E}_I^4) \vspace{1ex} \\
    \tilde{\mathbf{Q}}_v=CA(\mathbf{Q}_{v}^{\prime}, \mathbf{E}_{f}) + \mathbf{Q}_{v}^{\prime}
\end{array},
\right.
\end{equation}
where $Linear$ is the linear projection operation and $CA$ represents cross-attention operation.
% learnable prompts, 
After that, we further interact the enhanced video-level queries $\tilde{\mathbf{Q}}_v$ and $\mathbf{E}_I^4$ via element-wise multiplication, generating learnable embeddings $\mathbf{E}_L \in \mathbb{R}^{N_v \times c}$, which can be used to replace SAM’s original sparse prompt embeddings.
% mask decoder
Embarking on this, the mask decoder utilizes both learnable embeddings $\mathbf{E}_L$ and the image embeddings $\mathbf{E}_I$ to generate the predicted map $\mathbf{P}$ for the current frame. %, as shown in Fig. \ref{fig:model_structure}.

% tab:quantitative
\begin{table*}
    \centering
    
    \setlength{\tabcolsep}{1.5 mm}
    \begin{tabular}{l|c|cccc|cccc|cccc}
      \toprule
      \multirow{2}{*}{Methods} & \multirow{2}{*}{Referece}  &
      \multicolumn{4}{c|}{RDVS~\cite{mou2024salient}} & \multicolumn{4}{c|}{ViDSOD-100~\cite{lin2024vidsod}} & \multicolumn{4}{c}{DViSal~\cite{li2023dvsod}} \\ \cline{3-14}
      &  & $E_\xi \uparrow$ & $S_\alpha \uparrow$ & $F_\beta \uparrow$ & $M \downarrow$ & $E_\xi \uparrow$ & $S_\alpha \uparrow$ & $F_\beta \uparrow$ & $M \downarrow$ & $E_\xi \uparrow$ & $S_\alpha \uparrow$ & $F_\beta \uparrow$ & $M \downarrow$ \\
      \midrule
  
      HRTransNet  & TCSVT'22
                  & 0.725  & 0.671  & 0.445  & 0.076
                  & 0.745  & 0.686  & 0.531  & 0.099
                  & 0.745  & 0.685  & 0.531  & 0.099 \\
      PICRNet     & MM'23
                  & 0.743  & 0.728  & 0.535  & 0.074
                  & 0.873  & 0.830  & 0.738  & 0.038
                  & 0.715  & 0.670  & 0.568  & 0.147 \\
      DVSOD       & NeurIPS'23
                  & 0.748   & 0.587 & 0.452  & 0.070
                  & 0.783  & 0.702  & 0.568  & 0.083
                  & 0.807  & 0.729  & 0.610  & 0.113 \\
      LSTA        & PR'24
                  & 0.746   & 0.650 & 0.484  & 0.069
                  & 0.757  & 0.671  & 0.565  & 0.086
                  & 0.848  & 0.700  & 0.640  & 0.082 \\
      DPA         & CVPR'24
                  & 0.675  & 0.666  & 0.445  & 0.096
                  & 0.848  & 0.817  & 0.715  & 0.051
                  & 0.796  & 0.724  & 0.635  & 0.102 \\
      DCTNet+     & TIP'24
                  & 0.909   & 0.876 & 0.794  & 0.029
                  & 0.901  & 0.876  & 0.809  & 0.030
                  & 0.828  & 0.767  & 0.689  & 0.095 \\
      ATFNet      & IJCV'24
                  & 0.732   & 0.713 & 0.491  & 0.074
                  & 0.901  & 0.875  & 0.813  & 0.027
                  & 0.795  & 0.724  & 0.622  & 0.111 \\
      MDSAM       & MM'24
                  & 0.813  & 0.791  & 0.647  & 0.056
                  & 0.909  & 0.877  & 0.815  & 0.026
                  & 0.856  & 0.796  & 0.715  & 0.071 \\
      SAM2-UNet   & Arxiv'24
                  & 0.888  & 0.843  & 0.765  & 0.035
                  & 0.907  & 0.891  & 0.829  & 0.025
                  & 0.856  & 0.778  & 0.747  & 0.064 \\
      MFENet      & ICASSP'25
                  & -      & 0.794  & 0.700  & 0.049
                  & -      & 0.831  & 0.763  & 0.040
                  & -      & 0.760  & 0.717  & 0.080 \\
      KAN-SAM     & ICME'25
                  & 0.888  & 0.854  & 0.791  & 0.028
                  & 0.912  & 0.892  & 0.846  & 0.025
                  & 0.885  & 0.835  & 0.783  & 0.052 \\
      \textbf{Ours} & -
                  & \textbf{0.913} & \textbf{0.879} & \textbf{0.827} & \textbf{0.026}
                  & \textbf{0.918} & \textbf{0.894} & \textbf{0.868} & \textbf{0.020}
                  & \textbf{0.914} & \textbf{0.840} & \textbf{0.818} & \textbf{0.046} \\
      \bottomrule
    \end{tabular}
    \caption{Quantitative comparison results with the state-of-the-art RGB-D video salient object detection models on three representative datasets.}
    \label{tab:quantitative}
\end{table*}

\begin{table*}
    \centering
    
    \setlength{\tabcolsep}{1.8 mm}
    \begin{tabular}{l|cc|cccc}
      \toprule
      Methods & Trainable/Total (M) & Memory (G) & $E_\xi \uparrow$ & $S_\alpha \uparrow$ & $F_\beta \uparrow$ & $M \downarrow$ \\
      \midrule
  
      w/o depth projector                 & - & 20.3 & 0.899 & 0.870 & 0.808 & 0.023 \\
      w/o parallel (sequential adapter)   & 17.4/236.0 & 91.9 & 0.860 & 0.830 & 0.778 & 0.028 \\
      w/o parallel (LoRA)                 & 56.0/274.6 & 95.0 & 0.889 & 0.877 & 0.824 & 0.027 \\
      w/o multi-modal                     & - & 17.9 & 0.876 & 0.853 & 0.782 & 0.029 \\
      \textbf{Ours}                & \textbf{19.2/237.9} & \textbf{21.0} & \textbf{0.913} & \textbf{0.879} & \textbf{0.827} & \textbf{0.026} \\
      \bottomrule
  \end{tabular}
  \caption{Ablation studies of our Parallel Adapter-based Multi-modal Image Encoder (PAMIE).}
  \label{tab:A_parallel_adapter}
\end{table*}

% update
Different from the static frame-level queries, video-level queries are iteratively updated via an update mechanism, which effectively captures the temporal dependencies among different frames.
% 组成部分,SAM2, linear and Transformer decoder (cross attention + self attention + FFN)
The update mechanism comprises three components, including a memory encoder adopted from SAM2, a linear projection, and a transformer decoder module consisting of a cross-attention, a self-attention, and a feed-forward network (FFN).
% memory encoder
Concretely, given the image embeddings $\mathbf{E}_{I,t}$ and corresponding prediction map $\mathbf{P}_t$ at frame $t$, we first extract memory features as follows,
% eq:update_memory_feature
\begin{equation}\label{eq:update_memory_feature}
    \mathbf{F}_{m}=Linear(ME(\mathbf{E}_{I,t}, \mathbf{P}_t)),
\end{equation}
where $ME$ is the memory encoder. These features are then used for the temporal update stage, namely
% eq:update
\begin{equation}\label{eq:update}
    \mathbf{Q}_{v,t+1} = \mathbf{Q}_{v,t} + FFN(SA(CA(\mathbf{Q}_{v,t},\mathbf{F}_{m} ))),
\end{equation}
where $FFN$ and $SA$ represent the FFN and self-attention, respectively. This iterative update mechanism refines the video-level queries for the subsequent frame, which effectively builds the temporal dependencies. % are effectively captured.

% % 再次集中阐述该模块的作用
% Overall, the QTM module eliminates the need for handcrafted prompts and mitigates the computational burden of large memory banks, enabling effective temporal modeling for RGB-D video salient object detection.

\begin{figure*}
    \centering
    \includegraphics[width=0.80\linewidth]{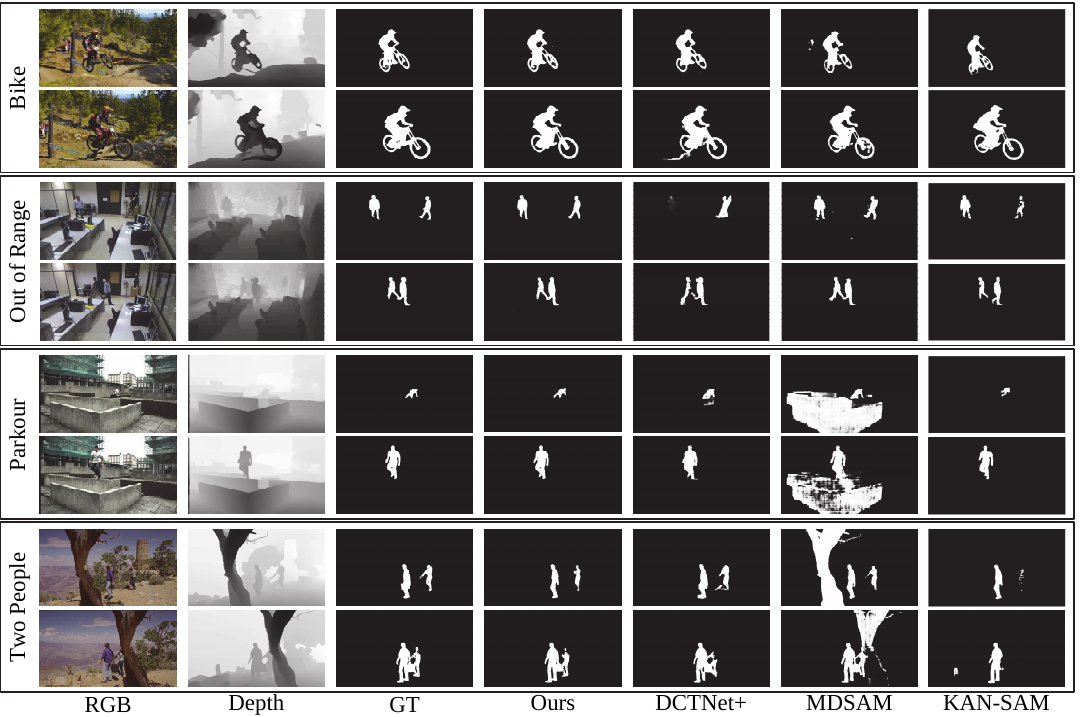}
    \caption{Qualitative comparison with the state-of-the-art RGB-D video salient object detection models on RDVS dataset.}
    \label{fig:visual_comparison}
\end{figure*}

% Loss
\subsection{Loss Function}
To effectively train our model, we attempt to supervise the generation of both image embeddings and learnable embeddings. Concretely, our loss function consists of the final prediction loss $\mathcal{L}_{\text{pred}}$ and the intermediate supervision loss $\mathcal{L}_{\text{inter}}$. The loss function $\mathcal{L}_{\text{total}}$ can be written as
% eq:loss
\begin{equation}\label{eq:loss}
    \mathcal{L}_{\text{total}} = \mathcal{L}_{\text{pred}} + \alpha \cdot \mathcal{L}_{\text{inter}},
\end{equation}
where both $\mathcal{L}_{\text{pred}}$ and $\mathcal{L}_{\text{inter}}$ are computed by using the binary cross entropy (BCE) loss~\cite{de2005tutorial}, and $\alpha$ is the weight of the intermediate loss. The final prediction loss $\mathcal{L}_{\text{pred}}$ computes the differences between the ground-truth $\mathbf{GT}$ and the final prediction result $\mathbf{P}$. The intermediate supervision loss $\mathcal{L}_{\text{inter}}$ is computed by comparing $\mathbf{GT}$ with intermediate prediction results $\tilde{\mathbf{P}}$.

\section{Experiments}

% 该部分参考《Multi-Scale and Detail-Enhanced Segment Anything Model for Salient Object Detection》，《Lightweight Multi-Frequency Enhancement Network for RGB-D Video Salient Object Detection》，他们均未详细介绍evaluation metrics. 在DViSal数据集的论文中有介绍，看着篇幅有点长，在evaluation metric站部分提供了引用
\subsection{Experiment Settings}

\label{sec:ex_Datasets}
\noindent{\bf Datasets.}
To comprehensively evaluate our model, we train and test our proposed method on three newly RGB-D VSOD datasets, namely RDVS~\cite{mou2024salient} (4087 frames, 57 videos), ViDSOD-100~\cite{lin2024vidsod} (9362 frames, 100 videos) and DViSal~\cite{li2023dvsod} (7117 frames, 237 videos). Note that, for a fair comparison, only the labeled frames in the DViSal dataset are used. %, under the same settings as other two datasets.

\label{sec:ex_evaluation_metrics}
\noindent{\bf Evaluation Metrics.}
Following the previous works~\cite{li2023dvsod}, we adopt four widely-used metrics to evaluate the model performance, \emph{i.e.}, E-measure ($E_\xi$)~\cite{fan2018enhanced}, S-measure ($S_\alpha$)~\cite{fan2017structure}, F-measure ($F_\beta$)~\cite{achanta2009frequency}, and mean absolute error (MAE or $M$)~\cite{borji2015salient}. \textit{The lower the MAE, the better. For other metrics, the higher score is better.}

\label{sec:ex_implementation_details}
\noindent{\bf Implementation Details.}
% SAM相关：1）采用SAM2.1 large；2）encoder冻结
Our SAM-DAQ is built on the large-scale configuration of SAM2 (SAM-L), with extensions for RGB-D VSOD.
During the training stage, the SAM encoder parameters are frozen, and the spatial resolution of input images is resized to $1024 \times 1024$.
% 视频处理策略，每个视频在每轮训练仅用10帧，为了获得长视频处理能力，这10帧一起送入网络
To balance training efficiency with long-term memory modeling, we randomly sample 10 frames per video in each epoch and feed them into the network simultaneously. We adopt AdamW~\cite{loshchilov2017decoupled} as the optimizer, where the learning rate and the weight decay are set to 0.0001 and 0.05, respectively. The batch size is set to 1, and the number of training iterations is 2000.
% 显存优势（3090）与训练效率优势（3小时）
Thanks to our efficient DPA, our SAM-DAQ can be trained on a single RTX-3090 (24 GB) GPU within 3 hours.

\begin{table}
    \centering
    
    \setlength{\tabcolsep}{2.2 mm}
    \begin{tabular}{l|cccc}
      \toprule
      Strategies & $E_\xi \uparrow$ & $S_\alpha \uparrow$ & $F_\beta \uparrow$ & $M \downarrow$ \\
      \midrule
      
      \textbf{sparse only (Ours)} & \textbf{0.913} & \textbf{0.879} & \textbf{0.827} & \textbf{0.026} \\
      dense only         & 0.875 & 0.856 & 0.783 & 0.032 \\
      both               & 0.862 & 0.839 & 0.763 & 0.033 \\
      \bottomrule
    \end{tabular}
    \caption{Ablation studies of our learnable embeddings generation strategy.}
    \label{tab:A_prompt_strategy}
\end{table}

\subsection{Comparison with the State-of-the-art Methods}

% sota
We compare SAM-DAQ with 11 state-of-the-art RGB-D VSOD models, including HRTransNet~\cite{tang2022hrtransnet}, PICRNet~\cite{cong2023point}, DVSOD~\cite{li2023dvsod}, LSTA~\cite{li2024efficient}, DPA~\cite{cho2024dual}, DCTNet+~\cite{mou2024salient}, ATFNet~\cite{lin2024vidsod}, MDSAM~\cite{gao2024multi}, SAM2-UNet~\cite{xiong2024sam2}, MFENet~\cite{suolang2025lightweight}, and KAN-SAM~\cite{li2025kansam}.

% quantitative
The quantitative results on three RGB-D VSOD datasets are shown in Table~\ref{tab:quantitative}. It can be observed that our SAM-DAQ achieves the best performance when compared with the cutting-edge methods. %Particularly, we achieve average improvements of 2.4\% in terms of E-measure, 2.1\% in terms of S-measure, 6.2\% in terms of F-measure, and 0.013 in terms of MAE when compared with the second-best results.
To demonstrate that the performance gain of our method is not solely due to the SAM2 backbone, we make comparisons with various SAM-based baselines. Specifically, compared with the KAN-SAM, we achieve average improvements of 1.5\%, 1.0\%, 2.4\% and 0.003 in terms of E-measure, S-measure, F-measure and MAE, respectively.
The above quantitative comparison results firmly demonstrate the effectiveness and superiority of our method.

% qualitative
The qualitative comparison is presented in Fig.~\ref{fig:visual_comparison}, where we compare our model with DCTNet+, MDSAM, and KAN-SAM on four representative scenes (``bike'', ``out of range'', ``parkour'' and ``two people'') from the RDVS benchmark.
% MDSAM没有利用时间信息，尽管可以完整的分割出物体，但分出得物体不一定是最显著的
We observe that though MDSAM and KAN-SAM can segment objects completely, they struggle to accurately detect the salient regions without temporal modeling.
% DCTNet+依赖光流法，对前景敏感，容易分错
DCTNet+ tends to highlight background regions, and the reason behind this can be attributed to that the quality of optical flow is highly sensitive to the foreground motion.
These visualization results further validate the consistent superiority of our SAM-DAQ.

\begin{figure}
    \centering
    \includegraphics[width=0.97\linewidth]{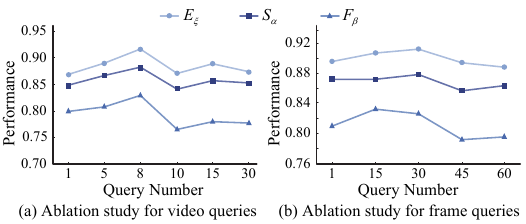}
    \caption{Ablation studies of different query numbers.}
    \label{fig:num_queries}
\end{figure}

\subsection{Ablation Studies}

\noindent{\bf Effect of PAMIE:}

% memory指标引入
To validate the effectiveness of our PAMIE, we conduct ablation studies and introduce trainable/total parameters (\emph{i.e.}, Trainable/Total) and GPU memory usage in the training stage (\emph{i.e.}, Memory) as additional metrics, as shown in Table~\ref{tab:A_parallel_adapter}.
% % 表格分析
% w/o depth projector
The results indicate that the E-measure drops from 91.3\% to 89.9\% when removing the depth projector (\emph{i.e.}, w/o depth projector), which highlights the necessity of addressing the inherent differences between RGB cues and depth cues. %the removal of the depth projector causes the E-measure to drop from 91.3\% to 89.9\%, highlighting the necessity of addressing the heterogeneous nature of RGB and depth modalities.
% w/o parallel
Replacing the parallel adapter with either a sequential adapter (\emph{i.e.}, w/o parallel (sequential adapter)) or LoRA (\emph{i.e.}, w/o parallel (LoRA)) increases memory usage significantly (91.9 GB and 95.0 GB, respectively), despite maintaining competitive accuracy. In stark contrast, our DPA achieves the best overall performance with only 21.0 GB of memory usage, confirming its superior trade-off between accuracy and memory efficiency.
% w/o multi-modal
% Compared with it, ``w/o multi-modal''solely remove the multi-modal fusion, the E-measure, S-measure and F-measure increase 1.6\%, 2.3\% and 0.4\%, respectively.
Besides, compared to the w/o multi-modal, we can see that our model still performs better than this variant. %, which demonstrates the effectiveness of the multi-modal fusion.% leads to a decline in metrics.
According to the above comparison results, we can confirm that our DPAs organized in a skip-connection way and explicit multi-modal fusion are crucial for effectively adapting SAM to the RGB-D VSOD task.
  
\begin{table}
    \centering
  
    \setlength{\tabcolsep}{3.2 mm}
    \begin{tabular}{l|cccc}
      \toprule
      Dimensions & $E_\xi \uparrow$ & $S_\alpha \uparrow$ & $F_\beta \uparrow$ & $M \downarrow$ \\
      \midrule
      
      32        & 0.889 & 0.863 & 0.795 & 0.023\\
      \textbf{64 (Ours)} & \textbf{0.913} & \textbf{0.879} & \textbf{0.827} & \textbf{0.026} \\
      128       & 0.874 & 0.842 & 0.779 & 0.030 \\
      256       & 0.880 & 0.863 & 0.799 & 0.023 \\
      \bottomrule
    \end{tabular}
    \caption{Ablation studies of query hidden dimension.}
    \label{tab:query_hidden_dimension}
\end{table}
  
\begin{table}
    \centering
  
    \setlength{\tabcolsep}{2.8 mm}
    \begin{tabular}{l|cccc}
      \toprule
      Methods & $E_\xi \uparrow$ & $S_\alpha \uparrow$ & $F_\beta \uparrow$ & $M \downarrow$ \\
      \midrule
      
      none            & 0.883 & 0.854 & 0.788 & 0.032 \\
      SAM2            & 0.853 & 0.829 & 0.796 & 0.028 \\
      multiply        & 0.895 & 0.862 & 0.804 & 0.027 \\
      \textbf{addition (Ours)} & \textbf{0.913} & \textbf{0.879} & \textbf{0.827} & \textbf{0.026} \\
      \bottomrule
    \end{tabular}
    \caption{Ablation studies of update mechanism in Query-Driven Temporal Memory (QTM) Module.}
    \label{tab:A_memory_module}
\end{table}

% ablation studies on QTM
\noindent{\bf Effect of QTM:}

% 介绍关于learnable embeddings生成策略的消融：替代sparse or dense embeddings
To validate the effectiveness of the learnable embeddings, we conduct ablation studies, as shown in Table~\ref{tab:A_prompt_strategy}. We can see that sparse-only variant achieves superior performance when compared with the dense-only or their combination.
% 可能的解释  after rubuttal
A plausible explanation is that the queries in QTM interact via token-wise attention rather than pixel-wise convolution, making them structurally analogous to sparse embeddings in SAM pretraining. This structural consistency enables more efficient adaptation.
% num_queries超参消融
Furthermore, in Fig.~\ref{fig:num_queries}, we present a detailed analysis of the impact of the number of frame-level queries and video-level queries on the performance of our SAM-DAQ.
% 结果分析
Specifically, reducing the number of video-level queries to 5 results in decreases of 2.6\%, 1.6\%, and 0.8\% in terms of E-measure, S-measure, and F-measure, respectively. Conversely, increasing the number of video-level queries to 10 introduces excessive background noise, which significantly degrades detection performance.
For frame-level queries, the results indicate that 30 queries provide sufficient guidance. %, and it will generate negligible improvements when further increasing the number of frame-level queries. % yield negligible improvements.
Overall, the best performance is achieved when the number of video-level queries and the frame-level queries are set to 8 and 30, respectively. %video queries and 30 frame queries are employed.
% query hidden dimension消融
Additionally, we conduct a comprehensive analysis of the query's hidden dimension. As shown in Table~\ref{tab:query_hidden_dimension}, when the hidden dimensions of the query are set to 64, we can obtain optimal results.

% update策略消融
We also analyze the temporal update mechanism in the QTM module, as shown in Table~\ref{tab:A_memory_module}.
% 结果分析
% none
From the results, we can see that directly removing the update phase (\emph{i.e.}, none), the performance drops 3\% in terms of E-measure, 2.5\% in terms of S-measure, 3.9\% in terms of F-measure, and increases 0.006 \% in terms of MAE, which proves the importance of the video-level queries update mechanism.
% SAM2 memory bank
We also compare SAM2’s original memory bank (\emph{i.e.}, SAM2) with our QTM, we can see that our QTM more effectively leverages temporal cues than traditional memory‑bank baselines.
% multiply
Additionally, replacing our addition update strategy with a multiply update strategy, the variant multiply presents minor performance drops. This validates the effectiveness of our design.% validating the design.

\begin{table}
    \centering
  
    \setlength{\tabcolsep}{2.6 mm}
    \begin{tabular}{ccc|cccc}
      \toprule
      \multicolumn{3}{c|}{Supervised Levels} & \multirow{2}{*}{$E_\xi \uparrow$} & \multirow{2}{*}{$S_\alpha \uparrow$} & \multirow{2}{*}{$F_\beta \uparrow$} & \multirow{2}{*}{$M \downarrow$} \\ \cline{1-3}
      $\mathbf{E}_I^2$ & $\mathbf{E}_I^3$ & $\mathbf{E}_I^4$ \\
      \midrule
      
      \Checkmark &            &            & 0.855 & 0.837 & 0.762 & 0.033 \\
                 & \Checkmark &            & 0.877 & 0.846 & 0.770 & 0.031 \\
                 &            & \Checkmark & \textbf{0.913} & \textbf{0.879} & \textbf{0.827} & \textbf{0.026} \\
                 & \Checkmark & \Checkmark & 0.858 & 0.833 & 0.741 & 0.036 \\
      \Checkmark & \Checkmark & \Checkmark & 0.845 & 0.824 & 0.749 & 0.032 \\
      \bottomrule
    \end{tabular}
    \caption{Ablation studies of our Intermediate Supervision.}
    \label{tab:A_auxiliary_supervision}
\end{table}

% intermediate supervision
\noindent{\bf Effect of Intermediate Supervision:}
To validate the effectiveness of our intermediate supervision, we conduct ablation studies, where we deploy supervision to different levels of image embeddings.
% 结果分析
As shown in Table~\ref{tab:A_auxiliary_supervision},  supervising only the highest-level embeddings $\mathbf{E}_I^4$ yields the best performance, while additional supervision on lower-level embeddings degrades the overall performance.

\section{Conclusion}
In this paper, we propose a novel RGB-D VSOD model, namely SAM-DAQ, which introduces the vision foundation model (\emph{i.e.} SAM2) for RGB-D VSOD.
To address the challenges of prompt dependency, high memory consumption, and computational costs, we deploy two key components, namely the PAMIE and the QTM module.
% PAMIE
The PAMIE module leverages a series of DPAs that are deployed in a skip-connection way to efficiently integrate RGB and depth features while fine-tuning the encoder under prompt-free conditions.
% QTM
Meanwhile, the QTM module unifies the temporal modeling and prompts into a learnable query way, eliminating the need for handcrafted prompts and mitigating the computational burden of large memory banks.
% 实验
Extensive experiments on three benchmark RGB-D VSOD datasets demonstrate the superiority of our SAM-DAQ framework over state-of-the-art methods. Our approach establishes a new paradigm for leveraging vision foundation models in the RGB-D video salient object detection task, offering an efficient and effective solution for the RGB-D VSOD task.
% % future work
% In our future work, we will further optimize the efficiency of SAM and explore its generalization to other related tasks.
% future work - after rebuttal
In future work, we will further optimize the query-based memory for multi-object video segmentation and explore adaptive query generation strategies.

\section{Acknowledgments}
This work was supported in part by the Zhejiang Province Key R\&D Project No. 2023C01046, in part by the National Natural Science Foundation of China (No. 62271180, 62471278, 62471285).

\bibliography{reference}

\begin{thebibliography}{54}
\providecommand{\natexlab}[1]{#1}

\bibitem[{Achanta et~al.(2009)Achanta, Hemami, Estrada, and Susstrunk}]{achanta2009frequency}
Achanta, R.; Hemami, S.; Estrada, F.; and Susstrunk, S. 2009.
\newblock Frequency-tuned salient region detection.
\newblock In \emph{2009 IEEE conference on computer vision and pattern recognition}, 1597--1604. IEEE.

\bibitem[{Ayzenberg, Giryes, and Greenspan(2024)}]{ayzenberg2024protosam}
Ayzenberg, L.; Giryes, R.; and Greenspan, H. 2024.
\newblock ProtoSAM: One-Shot Medical Image Segmentation With Foundational Models.
\newblock \emph{arXiv preprint arXiv:2407.07042}.

\bibitem[{Bao et~al.(2024)Bao, Zhou, Lu, Sun, Yin, Hu, Zhang, and Yan}]{bao2024quality}
Bao, L.; Zhou, X.; Lu, X.; Sun, Y.; Yin, H.; Hu, Z.; Zhang, J.; and Yan, C. 2024.
\newblock Quality-aware selective fusion network for VDT salient object detection.
\newblock \emph{IEEE Transactions on Image Processing}.

\bibitem[{Bao et~al.(2025)Bao, Zhou, Zheng, Cong, Yin, Zhang, and Yan}]{bao2025ifenet}
Bao, L.; Zhou, X.; Zheng, B.; Cong, R.; Yin, H.; Zhang, J.; and Yan, C. 2025.
\newblock IFENet: Interaction, Fusion, and Enhancement network for VDT Salient Object Detection.
\newblock \emph{IEEE Transactions on Image Processing}.

\bibitem[{Borji et~al.(2019)Borji, Cheng, Hou, Jiang, and Li}]{borji2019salient}
Borji, A.; Cheng, M.-M.; Hou, Q.; Jiang, H.; and Li, J. 2019.
\newblock Salient object detection: A survey.
\newblock \emph{Computational visual media}, 5(2): 117--150.

\bibitem[{Borji et~al.(2015)Borji, Cheng, Jiang, and Li}]{borji2015salient}
Borji, A.; Cheng, M.-M.; Jiang, H.; and Li, J. 2015.
\newblock Salient object detection: A benchmark.
\newblock \emph{IEEE transactions on image processing}, 24(12): 5706--5722.

\bibitem[{Carion et~al.(2020)Carion, Massa, Synnaeve, Usunier, Kirillov, and Zagoruyko}]{carion2020end}
Carion, N.; Massa, F.; Synnaeve, G.; Usunier, N.; Kirillov, A.; and Zagoruyko, S. 2020.
\newblock End-to-end object detection with transformers.
\newblock In \emph{European conference on computer vision}, 213--229. Springer.

\bibitem[{Cheng and Schwing(2022)}]{cheng2022xmem}
Cheng, H.~K.; and Schwing, A.~G. 2022.
\newblock Xmem: Long-term video object segmentation with an atkinson-shiffrin memory model.
\newblock In \emph{European Conference on Computer Vision}, 640--658. Springer.

\bibitem[{Cho et~al.(2024)Cho, Lee, Lee, Lee, Choi, Kim, and Lee}]{cho2024dual}
Cho, S.; Lee, M.; Lee, S.; Lee, D.; Choi, H.; Kim, I.-J.; and Lee, S. 2024.
\newblock Dual prototype attention for unsupervised video object segmentation.
\newblock In \emph{Proceedings of the IEEE/CVF Conference on Computer Vision and Pattern Recognition}, 19238--19247.

\bibitem[{Cong et~al.(2023)Cong, Liu, Zhang, Zhang, Zheng, Song, and Kwong}]{cong2023point}
Cong, R.; Liu, H.; Zhang, C.; Zhang, W.; Zheng, F.; Song, R.; and Kwong, S. 2023.
\newblock Point-aware interaction and cnn-induced refinement network for RGB-D salient object detection.
\newblock In \emph{Proceedings of the 31st ACM international conference on multimedia}, 406--416.

\bibitem[{De~Boer et~al.(2005)De~Boer, Kroese, Mannor, and Rubinstein}]{de2005tutorial}
De~Boer, P.-T.; Kroese, D.~P.; Mannor, S.; and Rubinstein, R.~Y. 2005.
\newblock A tutorial on the cross-entropy method.
\newblock \emph{Annals of operations research}, 134: 19--67.

\bibitem[{Deng et~al.(2024)Deng, Wu, Zeng, and Qin}]{deng2024memsam}
Deng, X.; Wu, H.; Zeng, R.; and Qin, J. 2024.
\newblock MemSAM: taming segment anything model for echocardiography video segmentation.
\newblock In \emph{Proceedings of the IEEE/CVF Conference on Computer Vision and Pattern Recognition}, 9622--9631.

\bibitem[{Diao et~al.(2024)Diao, Wan, Zhang, Jia, Lu, and Chen}]{diao2024unipt}
Diao, H.; Wan, B.; Zhang, Y.; Jia, X.; Lu, H.; and Chen, L. 2024.
\newblock Unipt: Universal parallel tuning for transfer learning with efficient parameter and memory.
\newblock In \emph{Proceedings of the IEEE/CVF Conference on Computer Vision and Pattern Recognition}, 28729--28740.

\bibitem[{Fan et~al.(2017)Fan, Cheng, Liu, Li, and Borji}]{fan2017structure}
Fan, D.-P.; Cheng, M.-M.; Liu, Y.; Li, T.; and Borji, A. 2017.
\newblock Structure-measure: A new way to evaluate foreground maps.
\newblock In \emph{Proceedings of the IEEE international conference on computer vision}, 4548--4557.

\bibitem[{Fan et~al.(2018)Fan, Gong, Cao, Ren, Cheng, and Borji}]{fan2018enhanced}
Fan, D.-P.; Gong, C.; Cao, Y.; Ren, B.; Cheng, M.-M.; and Borji, A. 2018.
\newblock Enhanced-alignment measure for binary foreground map evaluation.
\newblock \emph{arXiv preprint arXiv:1805.10421}.

\bibitem[{Fan, Wang, and Liang(2015)}]{fan2015improving}
Fan, D.~P.; Wang, J.; and Liang, X.~M. 2015.
\newblock Improving image retrieval using the context-aware saliency areas.
\newblock \emph{Applied Mechanics and Materials}, 734: 596--599.

\bibitem[{Fang et~al.(2024)Fang, Zhang, Zhou, and Zhang}]{fang2024learning}
Fang, H.; Zhang, T.; Zhou, X.; and Zhang, X. 2024.
\newblock Learning better video query with sam for video instance segmentation.
\newblock \emph{IEEE Transactions on Circuits and Systems for Video Technology}.

\bibitem[{Gao et~al.(2024)Gao, Zhang, Yan, and Lu}]{gao2024multi}
Gao, S.; Zhang, P.; Yan, T.; and Lu, H. 2024.
\newblock Multi-scale and detail-enhanced segment anything model for salient object detection.
\newblock In \emph{Proceedings of the 32nd ACM International Conference on Multimedia}, 9894--9903.

\bibitem[{Han et~al.(2017)Han, Chen, Liu, Yan, and Li}]{han2017cnns}
Han, J.; Chen, H.; Liu, N.; Yan, C.; and Li, X. 2017.
\newblock CNNs-based RGB-D saliency detection via cross-view transfer and multiview fusion.
\newblock \emph{IEEE transactions on cybernetics}, 48(11): 3171--3183.

\bibitem[{Hao et~al.(2024)Hao, Xiao, Luo, Guo, Wang, Shen, and Hu}]{hao2024primkd}
Hao, Z.; Xiao, Z.; Luo, Y.; Guo, J.; Wang, J.; Shen, L.; and Hu, H. 2024.
\newblock PrimKD: Primary Modality Guided Multimodal Fusion for RGB-D Semantic Segmentation.
\newblock In \emph{Proceedings of the 32nd ACM International Conference on Multimedia}, 1943--1951.

\bibitem[{Houlsby et~al.(2019)Houlsby, Giurgiu, Jastrzebski, Morrone, De~Laroussilhe, Gesmundo, Attariyan, and Gelly}]{houlsby2019parameter}
Houlsby, N.; Giurgiu, A.; Jastrzebski, S.; Morrone, B.; De~Laroussilhe, Q.; Gesmundo, A.; Attariyan, M.; and Gelly, S. 2019.
\newblock Parameter-efficient transfer learning for NLP.
\newblock In \emph{International conference on machine learning}, 2790--2799. PMLR.

\bibitem[{Hu et~al.(2022)Hu, Shen, Wallis, Allen-Zhu, Li, Wang, Wang, Chen et~al.}]{hu2022lora}
Hu, E.~J.; Shen, Y.; Wallis, P.; Allen-Zhu, Z.; Li, Y.; Wang, S.; Wang, L.; Chen, W.; et~al. 2022.
\newblock Lora: Low-rank adaptation of large language models.
\newblock \emph{ICLR}, 1(2): 3.

\bibitem[{Huang et~al.(2025)Huang, Wu, Zhou, Lin, Chen, Zhang, Xia, and Zhang}]{huang2025multi}
Huang, J.; Wu, Y.; Zhou, X.; Lin, J.; Chen, Z.; Zhang, G.; Xia, L.; and Zhang, J. 2025.
\newblock Multi-Scale Adaptive Prototype Transformer Network for Few-shot Strip Steel Surface Defect Segmentation.
\newblock \emph{IEEE Transactions on Instrumentation and Measurement}.

\bibitem[{Ji et~al.(2020)Ji, Zhang, Jie, Ma, and Wu}]{ji2020casnet}
Ji, Y.; Zhang, H.; Jie, Z.; Ma, L.; and Wu, Q.~J. 2020.
\newblock CASNet: A cross-attention siamese network for video salient object detection.
\newblock \emph{IEEE transactions on neural networks and learning systems}, 32(6): 2676--2690.

\bibitem[{Kirillov et~al.(2023)Kirillov, Mintun, Ravi, Mao, Rolland, Gustafson, Xiao, Whitehead, Berg, Lo et~al.}]{kirillov2023segment}
Kirillov, A.; Mintun, E.; Ravi, N.; Mao, H.; Rolland, C.; Gustafson, L.; Xiao, T.; Whitehead, S.; Berg, A.~C.; Lo, W.-Y.; et~al. 2023.
\newblock Segment anything.
\newblock In \emph{Proceedings of the IEEE/CVF international conference on computer vision}, 4015--4026.

\bibitem[{Li et~al.(2018)Li, Xie, Wei, Wang, and Lin}]{li2018flow}
Li, G.; Xie, Y.; Wei, T.; Wang, K.; and Lin, L. 2018.
\newblock Flow guided recurrent neural encoder for video salient object detection.
\newblock In \emph{Proceedings of the IEEE conference on computer vision and pattern recognition}, 3243--3252.

\bibitem[{Li et~al.(2023)Li, Ji, Wang, Li et~al.}]{li2023dvsod}
Li, J.; Ji, W.; Wang, S.; Li, W.; et~al. 2023.
\newblock Dvsod: Rgb-d video salient object detection.
\newblock \emph{Advances in Neural Information Processing Systems}, 36: 8774--8787.

\bibitem[{Li et~al.(2024)Li, Zhang, Yuan, Xiao, Lin, and Xu}]{li2024efficient}
Li, P.; Zhang, Y.; Yuan, L.; Xiao, H.; Lin, B.; and Xu, X. 2024.
\newblock Efficient long-short temporal attention network for unsupervised video object segmentation.
\newblock \emph{Pattern Recognition}, 146: 110078.

\bibitem[{Li et~al.(2025)Li, Hou, Ren, and Wu}]{li2025kansam}
Li, X.; Hou, R.; Ren, T.; and Wu, G. 2025.
\newblock KAN-SAM: Kolmogorov-Arnold Network Guided Segment Anything Model for RGB-T Salient Object Detection.
\newblock In \emph{2025 IEEE International Conference on Multimedia and Expo (ICME)}, 1--6. IEEE.

\bibitem[{Lin et~al.(2024)Lin, Zhu, Shen, Fu, Zhang, and Wang}]{lin2024vidsod}
Lin, J.; Zhu, L.; Shen, J.; Fu, H.; Zhang, Q.; and Wang, L. 2024.
\newblock ViDSOD-100: A New Dataset and a Baseline Model for RGB-D Video Salient Object Detection.
\newblock \emph{International Journal of Computer Vision}, 132(11): 5173--5191.

\bibitem[{Liu and Liu(2023)}]{liu2023part}
Liu, Z.-y.; and Liu, J.-w. 2023.
\newblock Part-aware attention correctness for video salient object detection.
\newblock \emph{Engineering Applications of Artificial Intelligence}, 119: 105733.

\bibitem[{Loshchilov and Hutter(2017)}]{loshchilov2017decoupled}
Loshchilov, I.; and Hutter, F. 2017.
\newblock Decoupled weight decay regularization.
\newblock \emph{arXiv preprint arXiv:1711.05101}.

\bibitem[{Mou et~al.(2024)Mou, Lu, He, Min, Fu, and Zhao}]{mou2024salient}
Mou, A.; Lu, Y.; He, J.; Min, D.; Fu, K.; and Zhao, Q. 2024.
\newblock Salient object detection in RGB-D videos.
\newblock \emph{IEEE Transactions on Image Processing}.

\bibitem[{Mousselly-Sergieh et~al.(2018)Mousselly-Sergieh, Botschen, Gurevych, and Roth}]{mousselly2018multimodal}
Mousselly-Sergieh, H.; Botschen, T.; Gurevych, I.; and Roth, S. 2018.
\newblock A multimodal translation-based approach for knowledge graph representation learning.
\newblock In \emph{Proceedings of the Seventh Joint Conference on Lexical and Computational Semantics}, 225--234.

\bibitem[{Oh et~al.(2019)Oh, Lee, Xu, and Kim}]{oh2019video}
Oh, S.~W.; Lee, J.-Y.; Xu, N.; and Kim, S.~J. 2019.
\newblock Video object segmentation using space-time memory networks.
\newblock In \emph{Proceedings of the IEEE/CVF international conference on computer vision}, 9226--9235.

\bibitem[{Qu et~al.(2017)Qu, He, Zhang, Tian, Tang, and Yang}]{qu2017rgbd}
Qu, L.; He, S.; Zhang, J.; Tian, J.; Tang, Y.; and Yang, Q. 2017.
\newblock RGBD salient object detection via deep fusion.
\newblock \emph{IEEE transactions on image processing}, 26(5): 2274--2285.

\bibitem[{Ravi et~al.(2024)Ravi, Gabeur, Hu, Hu, Ryali, Ma, Khedr, R{\"a}dle, Rolland, Gustafson et~al.}]{ravi2024sam}
Ravi, N.; Gabeur, V.; Hu, Y.-T.; Hu, R.; Ryali, C.; Ma, T.; Khedr, H.; R{\"a}dle, R.; Rolland, C.; Gustafson, L.; et~al. 2024.
\newblock Sam 2: Segment anything in images and videos.
\newblock \emph{arXiv preprint arXiv:2408.00714}.

\bibitem[{Ryali et~al.(2023)Ryali, Hu, Bolya, Wei, Fan, Huang, Aggarwal, Chowdhury, Poursaeed, Hoffman et~al.}]{ryali2023hiera}
Ryali, C.; Hu, Y.-T.; Bolya, D.; Wei, C.; Fan, H.; Huang, P.-Y.; Aggarwal, V.; Chowdhury, A.; Poursaeed, O.; Hoffman, J.; et~al. 2023.
\newblock Hiera: A hierarchical vision transformer without the bells-and-whistles.
\newblock In \emph{International conference on machine learning}, 29441--29454. PMLR.

\bibitem[{Singh, Verma, and Cheruku(2024)}]{singh2024dsfnet}
Singh, H.; Verma, M.; and Cheruku, R. 2024.
\newblock Dsfnet: video salient object detection using a novel lightweight deformable separable fusion network.
\newblock \emph{IEEE Transactions on Instrumentation and Measurement}.

\bibitem[{Suolang et~al.(2025)Suolang, He, Tsering, Fu, Li, and Zhao}]{suolang2025lightweight}
Suolang, D.; He, J.; Tsering, W.; Fu, K.; Li, X.; and Zhao, Q. 2025.
\newblock Lightweight Multi-Frequency Enhancement Network for RGB-D Video Salient Object Detection.
\newblock In \emph{ICASSP 2025-2025 IEEE International Conference on Acoustics, Speech and Signal Processing (ICASSP)}, 1--5. IEEE.

\bibitem[{Tang et~al.(2022)Tang, Liu, Tan, and He}]{tang2022hrtransnet}
Tang, B.; Liu, Z.; Tan, Y.; and He, Q. 2022.
\newblock HRTransNet: HRFormer-driven two-modality salient object detection.
\newblock \emph{IEEE Transactions on Circuits and Systems for Video Technology}, 33(2): 728--742.

\bibitem[{Wang et~al.(2023)Wang, Chen, Wu, Luo, Tang, Dai, Zhao, Xie, Yuan, and Jiang}]{wang2023look}
Wang, J.; Chen, D.; Wu, Z.; Luo, C.; Tang, C.; Dai, X.; Zhao, Y.; Xie, Y.; Yuan, L.; and Jiang, Y.-G. 2023.
\newblock Look before you match: Instance understanding matters in video object segmentation.
\newblock In \emph{Proceedings of the IEEE/CVF conference on computer vision and pattern recognition}, 2268--2278.

\bibitem[{Wang et~al.(2024)Wang, Lin, Li, Tu, and Luo}]{wang2024adapting}
Wang, K.; Lin, D.; Li, C.; Tu, Z.; and Luo, B. 2024.
\newblock Adapting Segment Anything Model to Multi-modal Salient Object Detection with Semantic Feature Fusion Guidance.
\newblock \emph{arXiv preprint arXiv:2408.15063}.

\bibitem[{Wang, Shen, and Shao(2017)}]{wang2017video}
Wang, W.; Shen, J.; and Shao, L. 2017.
\newblock Video salient object detection via fully convolutional networks.
\newblock \emph{IEEE Transactions on Image Processing}, 27(1): 38--49.

\bibitem[{Xie et~al.(2025)Xie, Tang, Yan, and Agam}]{xie2025rfmedsam}
Xie, B.; Tang, H.; Yan, Y.; and Agam, G. 2025.
\newblock RFMedSAM 2: Automatic Prompt Refinement for Enhanced Volumetric Medical Image Segmentation with SAM 2.
\newblock \emph{arXiv preprint arXiv:2502.02741}.

\bibitem[{Xiong et~al.(2024)Xiong, Wu, Tan, Li, Tang, Chen, Li, Ma, and Li}]{xiong2024sam2}
Xiong, X.; Wu, Z.; Tan, S.; Li, W.; Tang, F.; Chen, Y.; Li, S.; Ma, J.; and Li, G. 2024.
\newblock Sam2-unet: Segment anything 2 makes strong encoder for natural and medical image segmentation.
\newblock \emph{arXiv preprint arXiv:2408.08870}.

\bibitem[{Xu(2025)}]{xu2025dgsunet}
Xu, Y. 2025.
\newblock DGSUnet: An Improved Unet Model with DINO-Guided SAM2 for Multi-Scale Feature Collaboration.
\newblock \emph{arXiv preprint arXiv:2503.21187}.

\bibitem[{Yang et~al.(2024)Yang, Bi, Zhang, and Sun}]{yang2024sam}
Yang, S.; Bi, H.; Zhang, H.; and Sun, J. 2024.
\newblock SAM-UNet: Enhancing Zero-Shot Segmentation of SAM for Universal Medical Images.
\newblock \emph{arXiv preprint arXiv:2408.09886}.

\bibitem[{Yue et~al.(2024)Yue, Zhang, Zhang, Zhao, Lv, and Ma}]{yue2024sam}
Yue, J.; Zhang, R.; Zhang, Z.; Zhao, R.; Lv, W.; and Ma, J. 2024.
\newblock How SAM helps Unsupervised Video Object Segmentation?
\newblock In \emph{2024 International Joint Conference on Neural Networks (IJCNN)}, 1--9. IEEE.

\bibitem[{Zhang et~al.(2020)Zhang, Liu, Wang, Lei, Wang, and Lu}]{zhang2020non}
Zhang, P.; Liu, W.; Wang, D.; Lei, Y.; Wang, H.; and Lu, H. 2020.
\newblock Non-rigid object tracking via deep multi-scale spatial-temporal discriminative saliency maps.
\newblock \emph{Pattern Recognition}, 100: 107130.

\bibitem[{Zhang et~al.(2024)Zhang, Liu, Lin, Liao, and Li}]{zhang2024uv}
Zhang, X.; Liu, Y.; Lin, Y.; Liao, Q.; and Li, Y. 2024.
\newblock Uv-sam: Adapting segment anything model for urban village identification.
\newblock In \emph{Proceedings of the AAAI Conference on Artificial Intelligence}, volume~38, 22520--22528.

\bibitem[{Zhong et~al.(2024)Zhong, Tang, He, Fang, and Yuan}]{zhong2024convolution}
Zhong, Z.; Tang, Z.; He, T.; Fang, H.; and Yuan, C. 2024.
\newblock Convolution meets lora: Parameter efficient finetuning for segment anything model.
\newblock \emph{arXiv preprint arXiv:2401.17868}.

\bibitem[{Zhou et~al.(2023)Zhou, Cao, Gao, Ming, and Zhang}]{zhou2023sti}
Zhou, X.; Cao, W.; Gao, H.; Ming, Z.; and Zhang, J. 2023.
\newblock STI-Net: Spatiotemporal integration network for video saliency detection.
\newblock \emph{Information Sciences}, 628: 134--147.

\bibitem[{Zhou et~al.(2021)Zhou, Pei, Li, Wang, Zheng, and He}]{zhou2021saliency}
Zhou, Z.; Pei, W.; Li, X.; Wang, H.; Zheng, F.; and He, Z. 2021.
\newblock Saliency-associated object tracking.
\newblock In \emph{Proceedings of the IEEE/CVF international conference on computer vision}, 9866--9875.

\end{thebibliography}

\end{document}